\documentclass[runningheads]{llncs}

\AtBeginDocument{%
  \providecommand\BibTeX{{%
    \normalfont B\kern-0.5em{\scshape i\kern-0.25em b}\kern-0.8em\TeX}}}

\usepackage{microtype}
\usepackage{graphicx}
\usepackage{caption}
\usepackage{subcaption}
\usepackage{booktabs} 
\usepackage{tabularx}
\usepackage{multicol}
\usepackage{empheq}
\usepackage{amsmath}
\usepackage{todonotes}
\usepackage{xspace} 
\usepackage{multirow}
\usepackage[ruled,norelsize,linesnumbered]{algorithm2e}
\usepackage{algorithmic}
\usepackage{tikz}
\usepackage[
  separate-uncertainty = true,
  multi-part-units = repeat
]{siunitx}

\newcommand{\etc}{\emph{etc.}\xspace}

\newcommand{\ie}{\emph{i.e.,}\xspace}
\newcommand{\eg}{\emph{e.g.,}\xspace}

\newcommand{\cut}[1]{}

\usepackage{hyperref}

\newcommand{\figref}[1]{Figure \ref{#1}}
\newcommand{\secref}[1]{Section \ref{#1}}

\title{Why is the video analytics accuracy fluctuating, and what can we do about it?}

\begin{document}
\mainmatter
\def\ECCVSubNumber{19}   
\titlerunning{Why is the video analytics accuracy fluctuating?} 
\authorrunning{Accepted at ECCV AROW workshop'22\ \ \ \ \ \ \ \ \ \ \ \ \ \ \  \ \ \ \ \ \ \ \ \ \ \ \ \ \ Sibendu et. al.} 

\author{Sibendu Paul, Kunal Rao, Giuseppe Coviello, Murugan Sankaradas, \\
Oliver Po, Y. Charlie Hu, Srimat Chakradhar}

\author{Sibendu Paul\inst{1} \and
Kunal Rao\inst{2} \and
Giuseppe Coviello\inst{2} \and Murugan Sankaradas\inst{2} \and \\Oliver Po \inst{2} \and Y. Charlie Hu\inst{1} \and Srimat Chakradhar\inst{2}
\institute{Purdue University, West Lafayette, IN, USA
\email{\{paul90,ychu\}@purdue.edu}\\
 \and
NEC Laboratories America, Inc, New Jersey, USA\\
\email{\{kunal,giuseppe.coviello,murugs,oliver,chak\}@nec-labs.com}}}

\pagestyle{headings}

\maketitle
\vspace{-0.1in}
\section{Abstract}
\vspace{-0.1in}
It is a common practice to think of a video as a sequence of images (frames), and re-use deep neural network models that are trained only on images for similar analytics tasks on videos. 
In this paper, we show
that this ``leap of faith'' that deep learning models that work well
on images will also work well on videos is actually flawed. 
We show
that even when a video camera is viewing a scene that is not changing in any human-perceptible way,
and we control for external factors like video compression and environment (lighting),
the accuracy of video analytics application fluctuates noticeably. 
These fluctuations occur
because successive frames produced by the video camera may look similar visually, but are perceived quite differently by the video analytics applications. 
We observed that the root cause for these fluctuations is the dynamic camera parameter changes that a video camera
automatically makes in order to capture and produce a visually pleasing
video. The camera inadvertently acts as an ``unintentional adversary'' because
 these slight
changes in the image pixel values in consecutive frames, as we show,  have a noticeably adverse impact on the accuracy of insights from video analytics tasks that re-use image-trained deep learning models.
To address this inadvertent adversarial effect from the camera, we explore the use of transfer learning techniques to improve learning in video analytics tasks through the transfer of knowledge from learning on image analytics tasks. 
%
Our experiments with a number of different cameras, and a variety of different video analytics tasks, show that the inadvertent adversarial effect from the camera can be noticeably offset by quickly re-training the deep learning models using transfer learning. In particular, we show that our newly trained Yolov5 model reduces fluctuation in object detection across frames, which leads to better tracking of objects ($\sim$40\% fewer mistakes in tracking). 
Our paper 
also provides new
directions and techniques to mitigate the camera's adversarial effect
on deep learning
models used for video analytics applications~\footnote{This paper is accepted at \textbf{ECCV Adversarial Robustness In the Real World (AROW) workshop'22}}.

\vspace{-0.1in}
\section{Introduction}
\vspace{-0.1in}


Significant progress in machine learning and computer vision~\cite{imagenet,connell2013retail,wang2015deep,viso}, along with the explosive growth in Internet
of Things (IoT), edge computing, and high-bandwidth access networks
such as 5G~\cite{QUALCOMM-5G,CNET-5G}, have led to the wide adoption
of video analytics systems. These systems deploy cameras throughout the
world to support diverse applications in entertainment, health-care,
retail, automotive, transportation, home automation, safety, and
security market segments.  
The global video analytics market is estimated to grow from \$5
billion in 2020 to \$21 billion by 2027, at a CAGR of
22.70\%~\cite{allied-market-research}.

Video analytics systems rely on state of the art (SOTA) deep learning models~\cite{imagenet} to make sense of the content in the video streams. It is a common practice to think of a video as a sequence of images (frames), and re-use deep learning models that are trained only on images for video analytics tasks. Large, image datasets like  {COCO}~\cite{lin2014microsoft} have made it possible to train highly-accurate SOTA deep learning models~\cite{liu2016ssd,chiu2020mobilenet,sinha2019thin,bochkovskiy2020yolov4,glenn_jocher_2022_6222936,tan2020efficientdet} that detect a variety of objects in images. 
In this paper, we take a closer look at the use of popular deep neural network models trained on large image datasets for predictions in critical video analytics tasks. We consider video segments from two popular benchmark video datasets~\cite{canel2019scaling,fang2019locality}.  These videos contain cars or persons, and we used several SOTA deep neural network (DNN) models for object detection and face detection tasks to make sense of the content in the video streams. Also, these videos exhibit minimal activity (\ie cars or persons are not moving appreciably and hence, largely static). Since the scenes are mostly static, the ground truth (total number of cars or persons) does not change appreciably from frame to frame within each video. Yet, we observe that the accuracy of tasks like object detection or face detection unexpectedly fluctuate noticeably for consecutive frames, rather than more or less stay the same. Such unexpected, noticeable fluctuations occur across different camera models and across different camera vendors. 

Such detection fluctuations from frame to frame have an adverse impact on applications that use insights from object or face detection to perform higher-level tasks like tracking objects or recognizing people. Understanding the causes for these unexpected fluctuations in accuracy, and proposing methods to mitigate the impact of these fluctuations, are the main goals of this paper. 
%
We investigate the causes of the accuracy fluctuations
of these SOTA deep neural network models on largely static
scenes by carefully considering factors external and internal to a video camera. 
We examine the impact of external factors like the environmental conditions (lighting), video compression and motion in the scene, and internal factors like camera parameter settings in a video camera, on the fluctuations in performance of image-trained deep neural network models. Even after carefully controlling for these external and internal factors, the accuracy fluctuations persist, and our experiments show that another cause for these fluctuations is the dynamic camera parameter changes that a video camera
automatically makes in order to capture and produce a visually pleasing
video. The camera inadvertently acts as an ``unintentional adversary'' because
 these slight
changes in image pixel values in consecutive frames, as we show, have a noticeably adverse impact on the accuracy of insights from video analytics tasks that re-use image-trained deep learning models.
To address this inadvertent adversarial effect from the camera, we explore ways to mitigate this effect and propose the transfer of knowledge from learning on image analytics tasks to video analytics tasks.  

In this paper, we make the following key contributions:

\begin{itemize}
\item We take a closer look at the use of popular deep learning models that are trained on large image datasets for predictions in critical video analytics tasks, and show that the accuracy of tasks like object detection or face detection unexpectedly fluctuate noticeably for consecutive frames in a video; consecutive frames capture the same scene and have the same ground truth. We show that such unexpected, noticeable fluctuations occur across different camera models and across different camera vendors.

\item We investigate the root causes of the accuracy fluctuations
of these SOTA deep neural network models on largely static
scenes by carefully considering factors external and internal to a video camera. We show that a video camera inadvertently acts as an ``unintentional adversary'' when it automatically makes camera parameter changes in order to capture and produce a visually pleasing video. 

\item We draw implications of the unintentional adversarial effect on
the practical use of computer vision models
and propose a simple yet effective technique to transfer knowledge from learning on image analytics tasks to video analytics. Our newly trained Yolov5 model reduces fluctuation in object detection across frames, which leads to better performance on object tracking task ($\sim$ 40\% fewer mistakes in tracking). 

\end{itemize}




\vspace{-0.25in}
\section{Motivation}
\label{sec:motivation}
\vspace{-0.05in}
In this section, we consider video segments from two popular benchmark datasets. These videos contain cars or persons, and the videos exhibit minimal activity (\ie cars or persons are not moving appreciably and hence, largely static). Since the scenes are mostly static, the ground truth (total number of cars or persons) from frame to frame is also not changing much.
Yet, we observe that the accuracy of tasks like object detection or face detection unexpectedly fluctuate noticeably for consecutive frames. 
Such accuracy fluctuations from frame to frame have an adverse impact on applications that use insights from object or face detection to perform higher-level tasks like tracking objects or recognizing people.    



\vspace{-0.15in}
\subsection{Object detection in videos}
\label{sec:motivation-obj-detection}
\vspace{-0.05in}
\begin{figure}
    \centering
    \begin{subfigure}[t]{0.495\textwidth}
        \vspace{-1.0\baselineskip}
        \centering
        \includegraphics[width=1.05\textwidth]{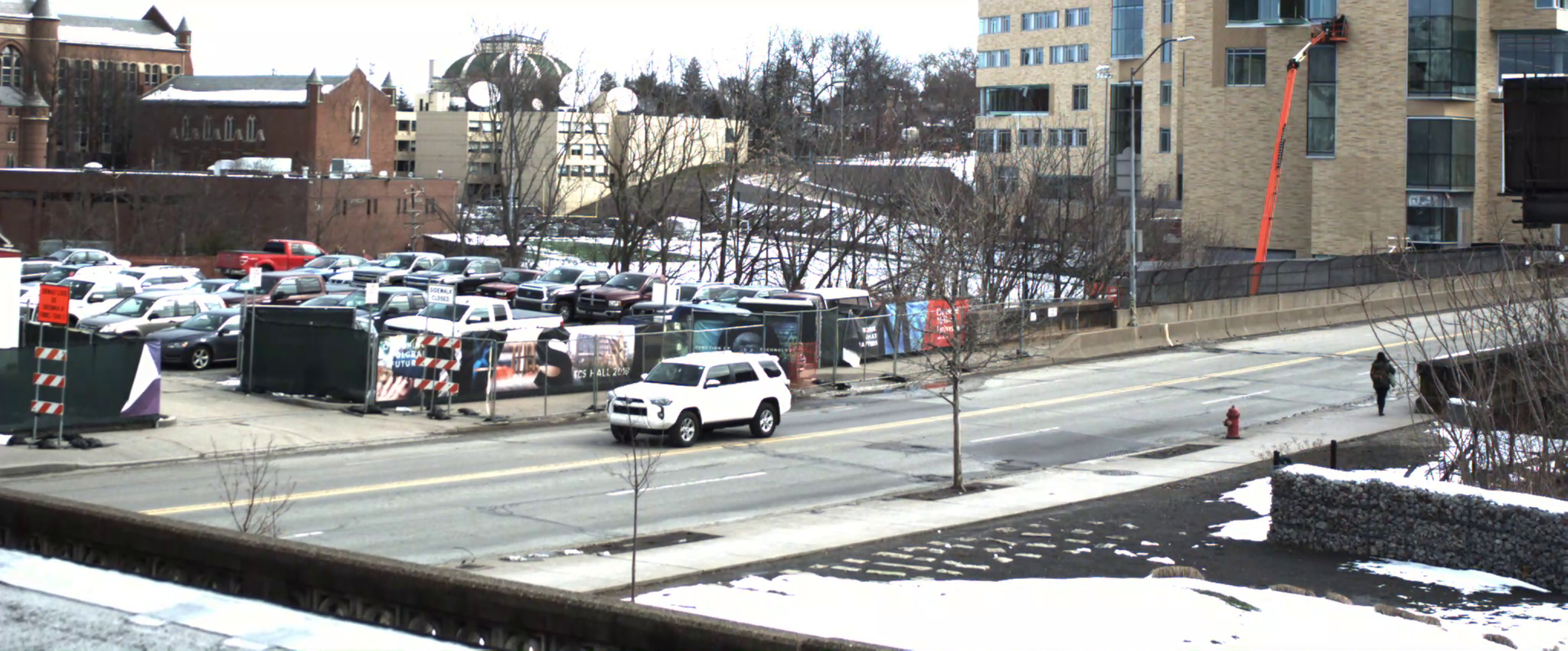}
        \caption{Roadway Dataset}
        \label{fig:roadway}
    \end{subfigure}
    \hfill
    \begin{subfigure}[t]{0.495\textwidth}
        \vspace{-1.0\baselineskip}
        \centering
        \includegraphics[width=0.8\textwidth]{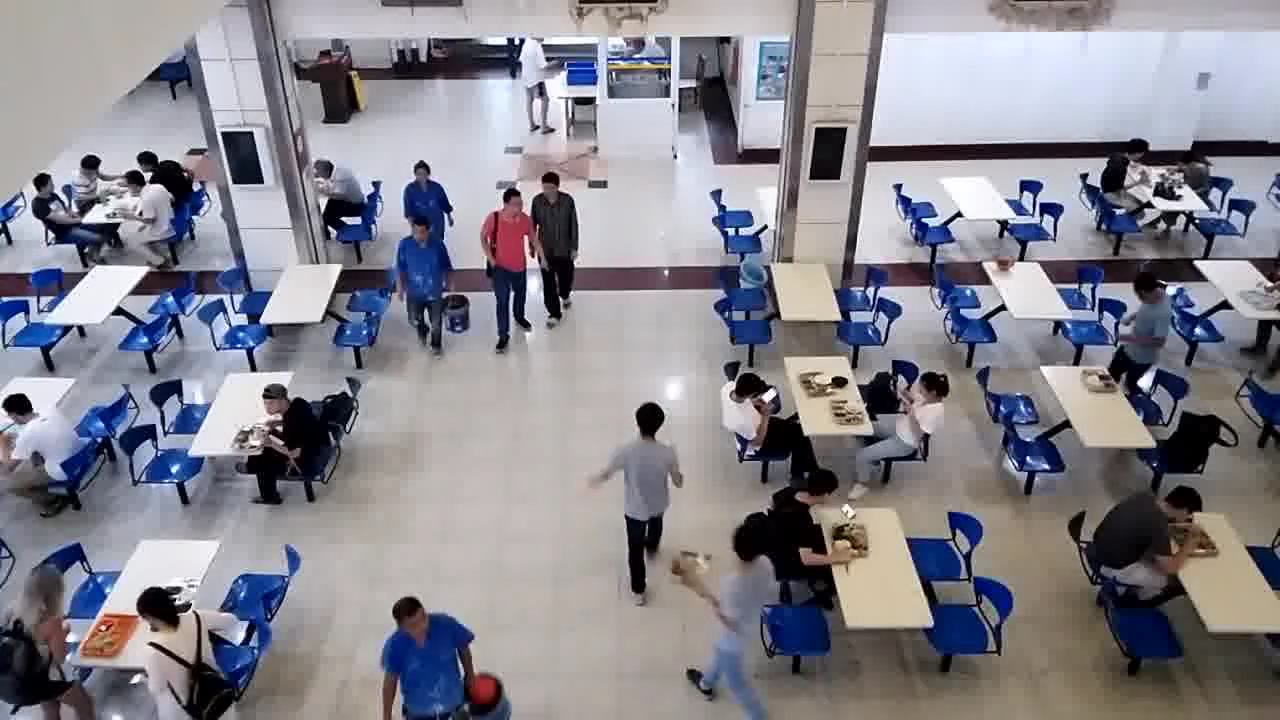}
        \caption{LSTN Dataset}
        \label{fig:facedet-prop}
    \end{subfigure}
    \vspace{-0.1in}
\caption{Sample frames from video datasets.}
\label{fig:prerecorded_dataset}
\vspace{-0.3in}
\end{figure}
One of the most common task in video analytics pipelines is object
detection. Detecting cars or people
is critical for many real-world applications like video surveillance,
retail, health care monitoring and intelligent transportation systems.

\begin{figure}
    \centering
    \begin{subfigure}[t]{0.47\textwidth}
        \vspace{-1.3\baselineskip}
        \centering
        \includegraphics[width=0.99\textwidth]{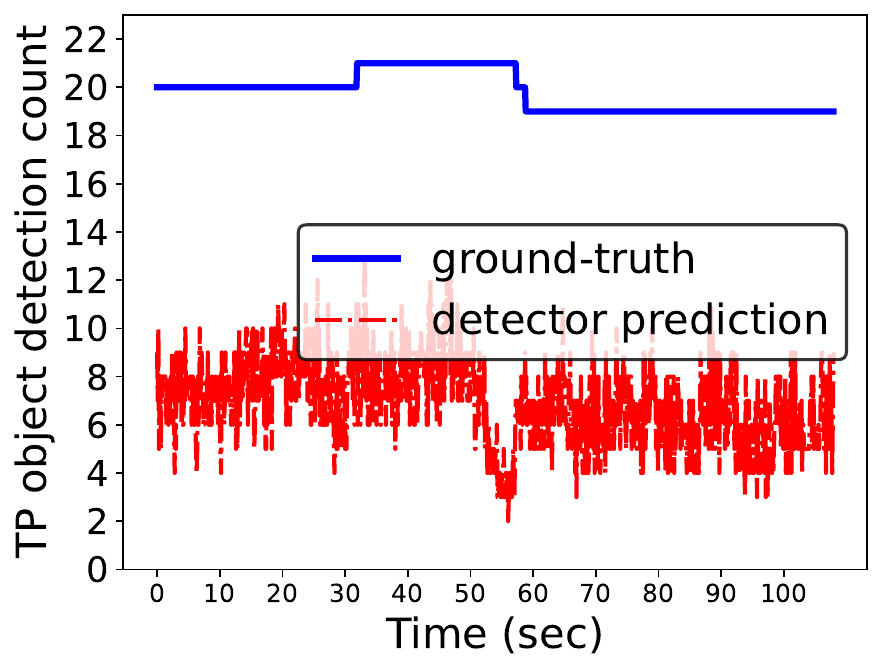}
        \caption{YOLOv5-small}
        \label{fig:yolov5s_roadway}
    \end{subfigure}
    \hfill
    \begin{subfigure}[t]{0.47\textwidth}
        \vspace{-1.3\baselineskip}
        \centering
        \includegraphics[width=0.99\textwidth]{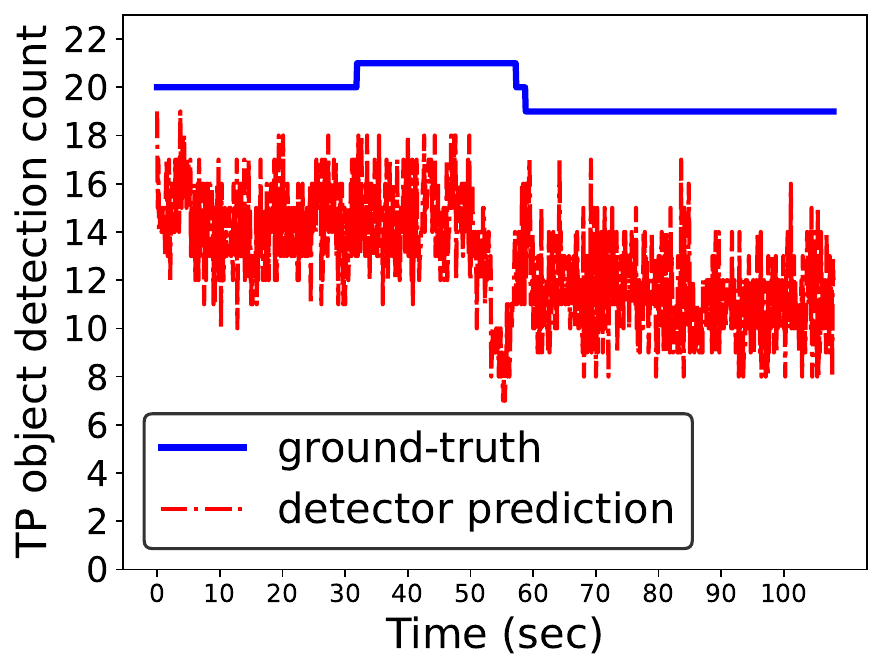}
        \caption{YOLOv5-large}
        \label{fig:yolov5l_roadway}
    \end{subfigure}
     \hfill
        \begin{subfigure}[t]{0.47\textwidth}
        \vskip 0pt
        \centering
        \includegraphics[width=0.99\textwidth]{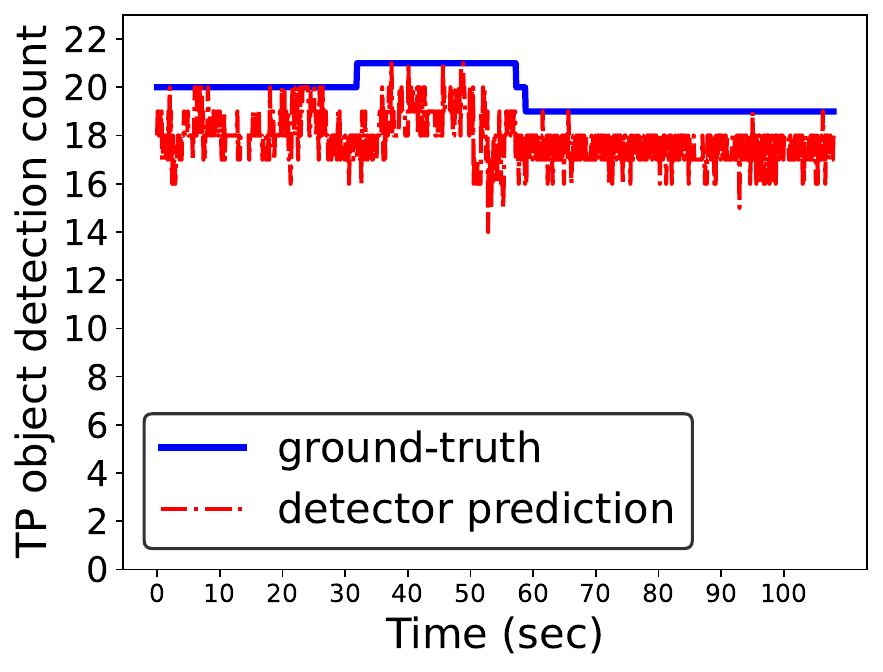}
        \caption{EfficientDet-v0}
        \label{fig:yolov5_v0}
    \end{subfigure}
         \hfill
        \begin{subfigure}[t]{0.47\textwidth}
        \vskip 0pt
        \centering
        \includegraphics[width=0.99\textwidth]{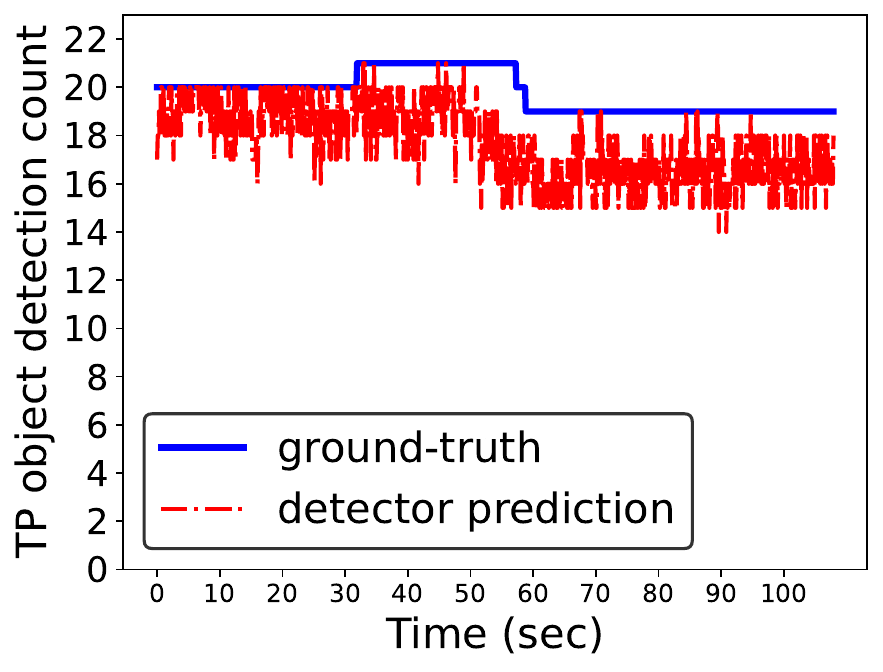}
        \caption{EfficientDet-v8}
        \label{fig:yolov5_v8}
    \end{subfigure}
    \vspace{-0.1in}
\caption{Performance of various object detection models on a segment of pre-recorded video from the Roadway dataset~\cite{canel2019scaling}.}
\label{fig:objdetector_prerecorded}
\vspace{-0.3in}
\end{figure}

Fig. \ref{fig:objdetector_prerecorded}
shows the performance of different state of the art and widely-used  object detectors like
{YOLO}v5-small and large variant ~\cite{glenn_jocher_2022_6222936}, Efficient{D}et-v0 and Efficient{D}et-v8~\cite{tan2020efficientdet} on video segments from the Roadway
dataset~\cite{canel2019scaling}. These videos have cars and people, but the activity is minimal, and scenes are largely static.  The
``ground truth" in the figures is shown in blue color, and it shows the total number of cars and people at different
times (i.e. frames) in the video. The ``detector prediction" waveform (shown in red color) shows the
number of cars and people actually detected by the deep learning model. 

Our experiments show that (a) for all the detectors we considered, the number of detected objects is lower than
the ground truth~\footnote{
we have 1-2 false positive detections for Yolov5 and efficient{D}et.}, and (b) more importantly,
even though the ground truth is not changing appreciably in consecutive frames, the
detections reported by the detectors vary noticeably, and (c) light-weight models like Yolov5-small or Yolov5-large exhibit a much higher range of detection fluctuations than the more heavier models like \emph{efficient{D}et}. However, the heavier deep learning models make inferences by consuming significantly more computing resources than the light-weight models.

\vspace{-0.15in}
\subsection{Face detection in videos}
\label{sec:motivation-face-detection}
Next, we investigate if accuracy fluctuation observed in object detection models
also occur in other image-trained AI models that are used in video analytics tasks. We chose AI models for face detection task, which is critical to 
many real-world applications \eg identifying a person of interest in airports, hospitals or arenas, and
authenticating individuals based on face-recognition for face-based
payments. 
\begin{figure}
    \centering
       \begin{subfigure}[t]{0.32\textwidth}
        \vspace{-0.6\baselineskip}
        \centering
        \includegraphics[width=0.99\textwidth]{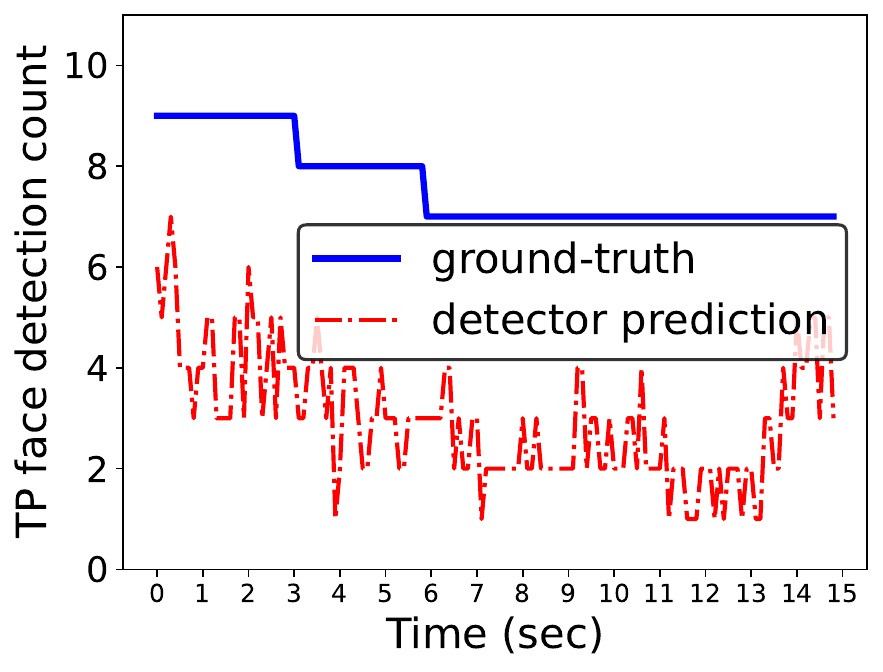}
        \caption{MTCNN}
        \label{fig:neoface_prerecord}
    \end{subfigure}
     \hfill
     \begin{subfigure}[t]{0.32\textwidth}
        \vspace{-0.6\baselineskip}
        \centering
        \includegraphics[width=0.99\textwidth]{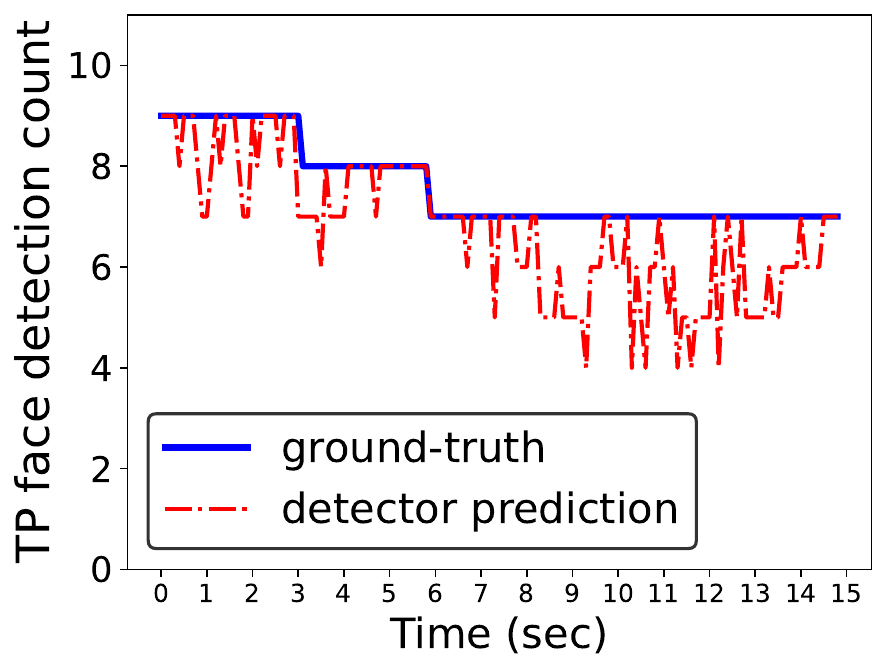}
        \caption{RetinaNet (Resnet-50)}
        \label{fig:retinaNet_prerecord}
    \end{subfigure}
    \hfill
    \begin{subfigure}[t]{0.32\textwidth}
        \vspace{-0.6\baselineskip}
        \centering
        \includegraphics[width=0.99\textwidth]{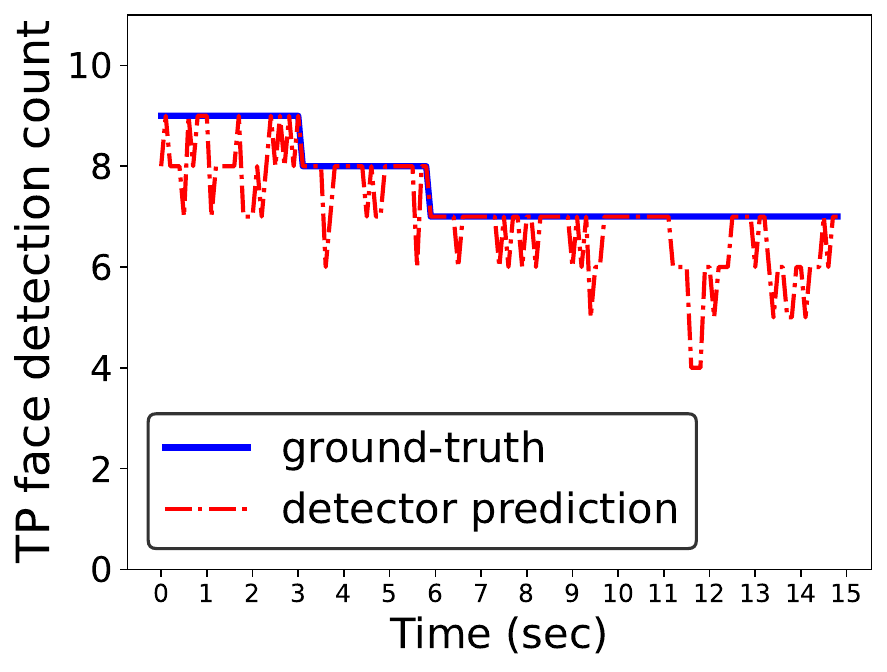}
        \caption{RetinaNet (Mobilenet)}
        \label{fig:mobilenetNet_prerecord}
    \end{subfigure}
    \vspace{-0.1in}
\caption{Performance of face detection models on videos from \emph{LSTN} video dataset.}
\label{fig:facedetector_prerecorded}
\vspace{-0.2in}
\end{figure}
Fig. \ref{fig:facedetector_prerecorded} shows the performance of three
well-known face detection AI models on videos from the \emph{LSTN} video
dataset~\cite{fang2019locality}.
Like the object detection case, we observe that (a) the number of faces detected by these models is typically lower than the ground truth, (b) more importantly, even though the ground truth barely changes, there is noticeable fluctuation in the number of detections in consecutive frames, and (c) the light-weight models like MTCNN~\cite{schroff2015facenet} exhibit a much higher range of detection fluctuations than the more heavier models like RetinaNet with resnet-50 and mobilenet backbone~\cite{deng2019retinaface}

\vspace{-0.2in}
\section{Analysis and control of external factors}
\label{sec:investigation}
\vspace{-0.05in}


The behavior of a DNN model is deterministic in the sense that if a frame is processed multiple times by the DNN model, then the DNN inferences are identical. In this section, we analyze three external factors that may be causing the unexpected accuracy fluctuations described in Section~\ref{sec:motivation}: 
\begin{itemize}
    \item Motion in the field of view of the camera affects the quality of the captured video (blurring of moving objects is likely).
    \item Lossy video compression methods like H.264 can also result in decoded frames whose quality can differ from the pre-compression frames. 
    \item Environmental conditions like lighting can also affect the quality of the frames processed by the DNNs. For example, flicker in fluorescent lighting can affect the quality of frames captured by the camera (most people cannot notice the flicker in fluorescent lights, which flicker at a rate of 120 cycles per second or 120 Hz; as we show later, flicker also contributes to fluctuations in the analytics accuracy of video analytics tasks).
\end{itemize}


\vspace{-0.15in}
\subsection{Control for motion}
\vspace{-0.05in}
It is difficult to systematically study the impact of motion on accuracy fluctuations by using videos from the datasets. Instead, as shown in Figure~\ref{fig:flicker_capture}, 
we set up a scene with 3D models of objects (\ie
persons and cars), and continuously observed the scene by using different IP cameras like \emph{AXIS Q1615}. A fluorescent light provides illumination for the scene. 
Figure~\ref{fig:flicker_capture} shows a
frame in the video stream captured by the IP camera under default camera parameter
settings. This setup easily eliminates the effect of motion on any observed accuracy fluctuations. Also, this set up makes it easy to study whether accuracy fluctuations are caused by only
certain camera models or fluctuations happen across different camera models from different vendors. 
\vspace{-0.15in}
\subsection{Analysis and control for video compression}
\label{subsec:video-compression}
\vspace{-0.05in}

By using static 3D models, we eliminated the effect of \textit{motion}. To understand the effect of video compression, we fetch frames directly from the camera instead of fetching a compressed video stream and decoding the stream to obtain frames that can be processed by a DNN model. 

\begin{figure}
    \centering
    \begin{subfigure}[t]{0.47\textwidth}
        \vspace{-0.7\baselineskip}
        \centering
        \includegraphics[width=0.99\textwidth]{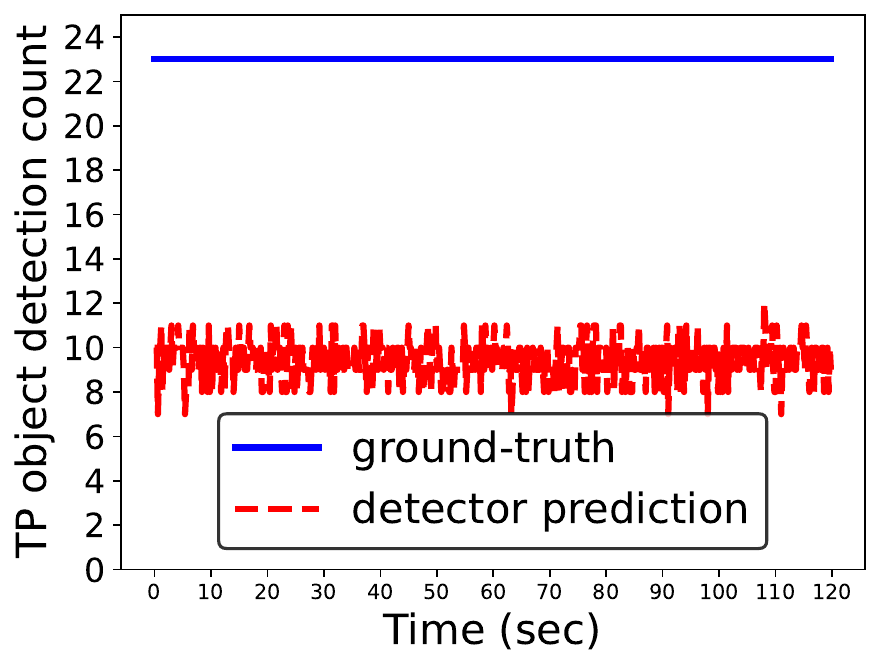}
        \caption{With Compression}
        \label{fig:yolov5_org}
    \end{subfigure}
    \hfill
    \begin{subfigure}[t]{0.47\textwidth}
        \vspace{-0.7\baselineskip}
        \centering
        \includegraphics[width=0.99\textwidth]{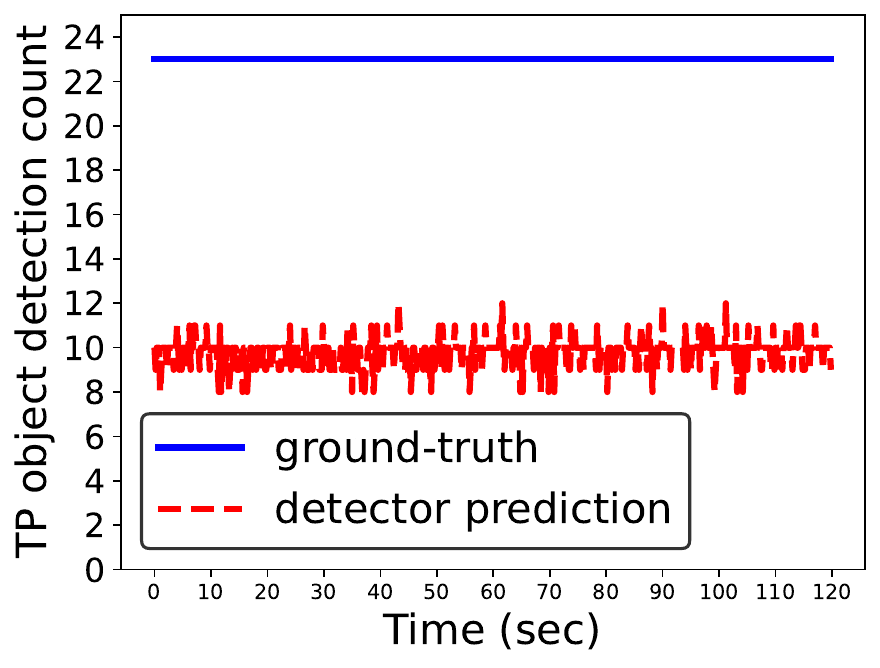}
        \caption{Without Compression}
        \label{fig:yolov5_nocomp}
    \end{subfigure}
    \vspace{-0.1in}
\caption{Effect of video compression on fluctuations in Yolov5 object detection counts (scene with 3D models)}
\label{fig:obj_compression}
\vspace{-0.3in}
\end{figure}

~\figref{fig:yolov5_org} and ~\figref{fig:yolov5_nocomp} show
the object detection counts with and without compression for the {YOLO}v5
model. We observe that eliminating compression reduces
detection fluctuation.
We also analyzed the detection counts with and without compression by using the \emph{t-test} for repeated measures~\cite{witte-witte-book}. Let $A$ be the sequence of true-positive object detection counts (per frame) for the experiment where video compression is used. Let $B$ be the sequence of true-positive object detection counts for the case when no compression is used. We compute a third sequence $D$ that is a sequence of pair-wise differences between the true-positive object count without compression and with compression (\ie $B-A$). 

Essentially, the use of difference scores converts a two-sample problem with $A$ and $B$ into a one-sample problem with $D$. Our null hypothesis states that compression has no effect on object detection counts (and we hypothesize a population mean of 0 for the difference scores). Our experiment with a sample size of 200 frames
showed that we can reject the null hypothesis at the 0.01 level of significance (99\% confidence), suggesting there is evidence that elimination of compression does reduce the accuracy fluctuations. Similar results were observed for sample sizes of 100 and 1000 frames. 

While \emph{t-test} measures the statistical difference between two distributions, it doesn't reflect on the fluctuations observed in repeated measures. 
We propose two metrics to quantify the observed fluctuations across a group of frames.
(1) \textit{F2} which is defined as $\frac{\|tp(i) - tp(i+1)\|}{mean(gt(i),gt(i+1))}$ for frame ${i}$, where $tp(i)$, $gt(i)$ are true-positive object detection count and ground-truth object count respectively on frame ${i}$ (on a moving window of 2 frames) and 
(2) \textit{F10} which is defined as $\frac{\|max(tp(i),...,tp(i+9)) - min(tp(i),...,tp(i+9))\|}{mean(gt(i),.., gt(i+9))}$ (on a moving window of 10 frames).

By eliminating video compression, the maximum variation in object count on static scene can be reduced from 17.4\% to 13.0\% (\textit{F2}) and from 19.0\% to 17.4\% (\textit{F10}).
Clearly, video compression is highly likely to have an adverse effect on accuracy fluctuations, and eliminating compression can improve results of deep learning models.  

\vspace{-0.1in}
\subsection{Analysis and control for flicker}
\label{subsec:video-compression}
\begin{figure}
    \centering
    \begin{subfigure}[t]{0.495\textwidth}
        \vspace{-1.3\baselineskip}
        \centering
        \includegraphics[width=0.99\textwidth]{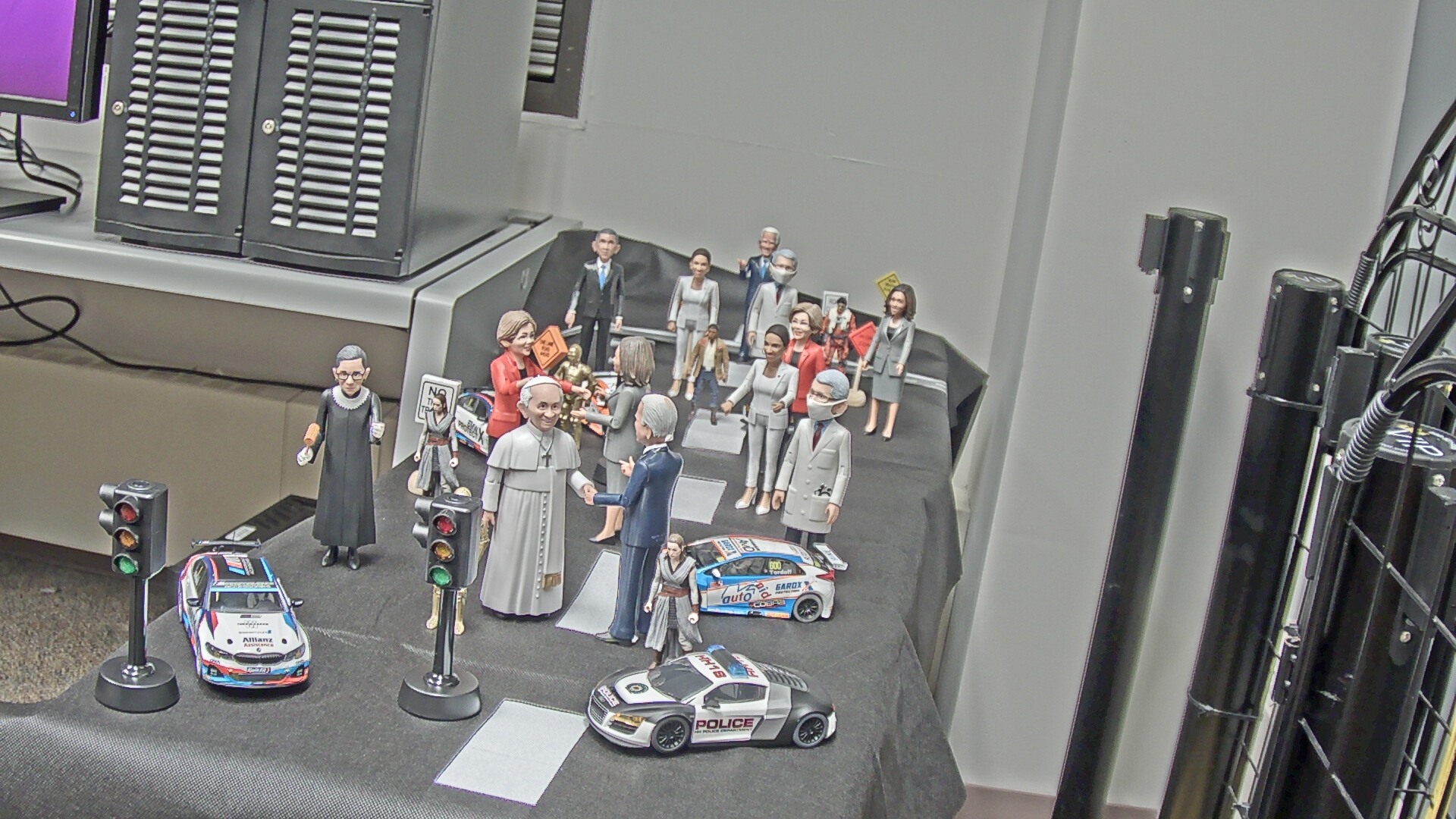}
        \caption{With flicker in lighting}
        \label{fig:flicker_capture}
    \end{subfigure}
    \hfill
    \begin{subfigure}[t]{0.495\textwidth}
        \vspace{-1.3\baselineskip}
        \centering
         \includegraphics[width=0.99\textwidth]{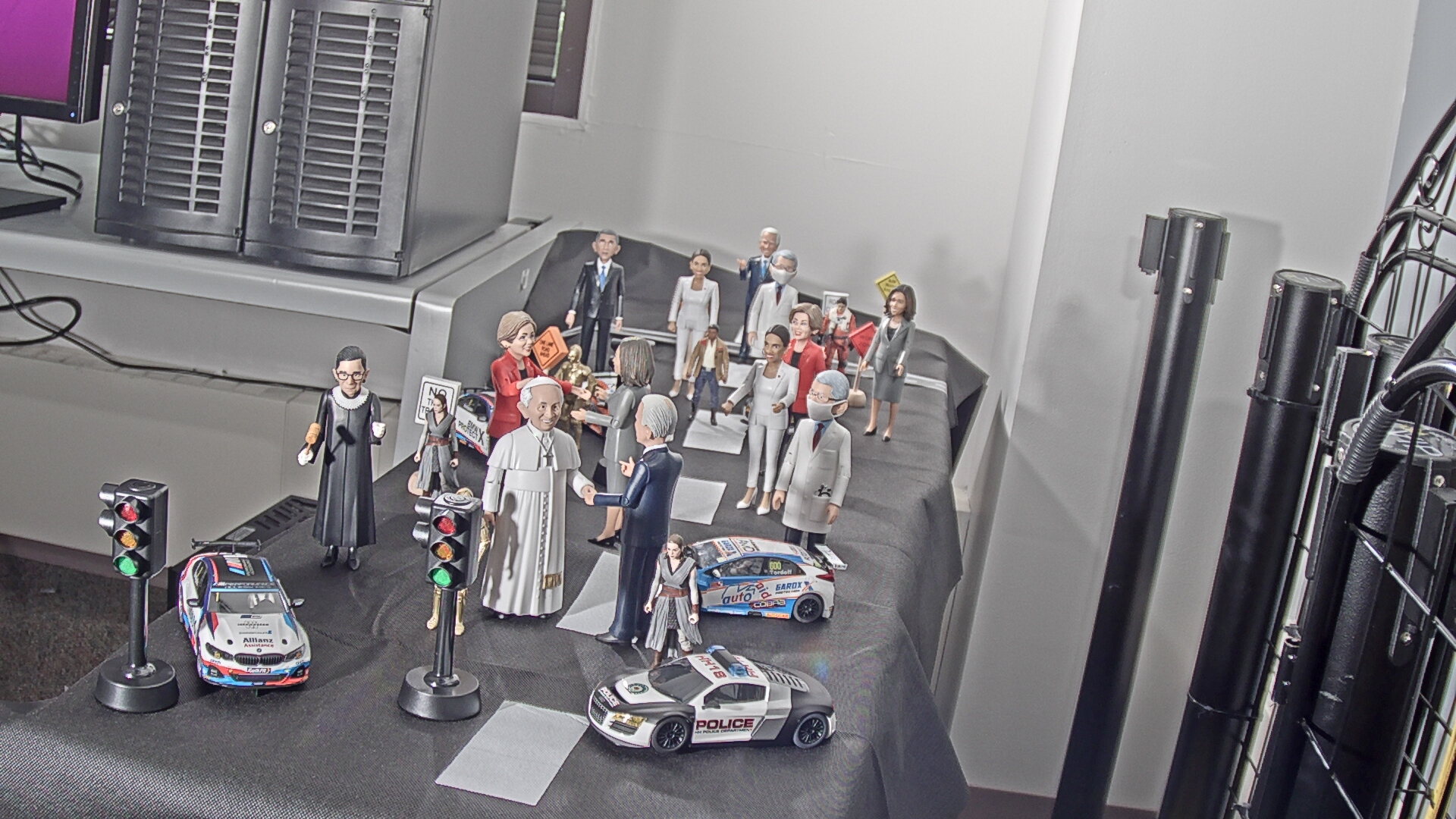}
        \caption{Without flicker in lighting}
        \label{fig:noflicker_capture}
    \end{subfigure}
    \vspace{-0.1in}
\caption{Scene with 3D models,  with and without flickering light.}
\label{fig:capture_flicker}
\vspace{-0.25in}
\end{figure}

\begin{figure}
    \centering
    \begin{subfigure}[t]{0.475\textwidth}
        \vspace{-0.6\baselineskip}
        \centering
        \includegraphics[width=0.99\textwidth]{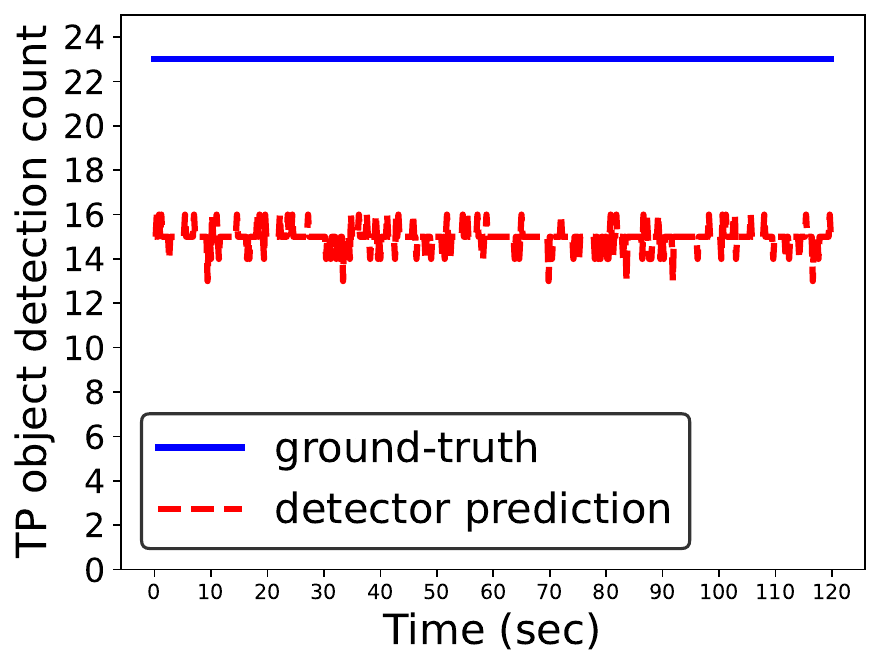}
        \caption{YOLOv5 detections}
        \label{fig:yolov5_flicker}
    \end{subfigure}
    \hfill
    \begin{subfigure}[t]{0.475\textwidth}
        \vspace{-0.6\baselineskip}
        \centering
        \includegraphics[width=0.99\textwidth]{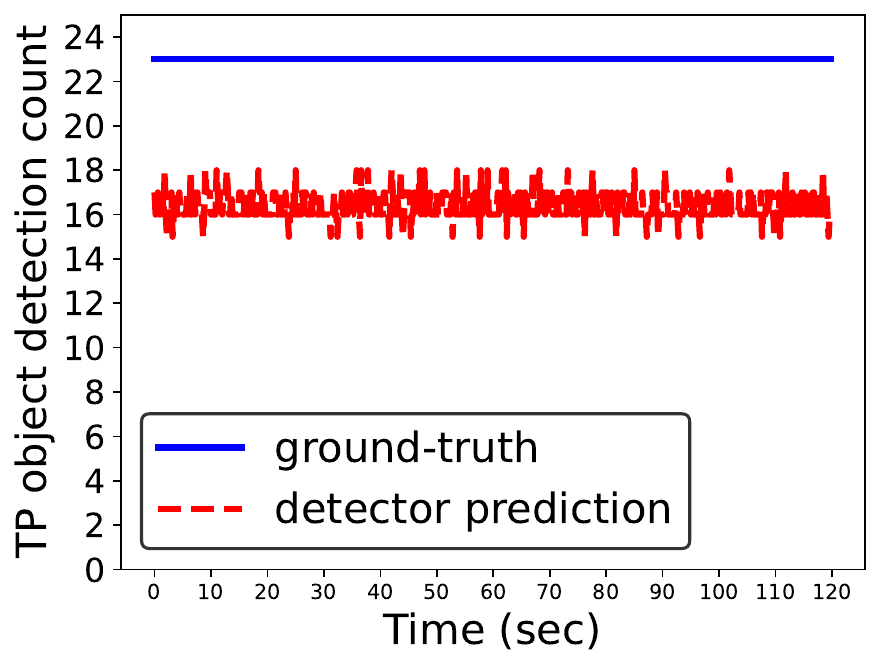}
        \caption{Efficient{D}et-v0 detections}
        \label{fig:efficientdet_flicker}
    \end{subfigure}
    \vspace{-0.1in}
\caption{Object detection counts when there is no motion, video compression or flickering light.}
\label{fig:obj_flicker}
\end{figure}

\vspace{-0.05in}
By using static 3D models, we eliminated the effect of \textit{motion}. We are also able to eliminate the adverse effect of video compression by fetching frames directly from the camera. We now analyze the effect of lighting. We set up an additional,  flicker-free light source to illuminate the scene with static 3D models. Figure~\ref{fig:capture_flicker} shows the 3D models scene with and without flickering light. 
~\figref{fig:yolov5_flicker} shows the fluctuation in detection counts when there is no motion, no video compression, and no flicker due to fluorescent light. 

Compared to Figure~\ref{fig:obj_compression} results with no compression (but with fluorescent lighting), the results in Figure~\ref{fig:yolov5_flicker} are highly likely to be an improvement. We compared the sequence of object detection counts  with and without fluorescent light (no video compression in both cases) using the t-test for repeated measures, and  easily rejected the null hypothesis that lighting makes no difference at a 0.01 level of significance (99\% confidence). Also, eliminating light flickering on top of motion and compression can reduce the maximum (\textit{F2}) and (\textit{F10}) variations from 13.0\% to 8.7\% and 17.4\% to 13.0\% respectively. Therefore, after eliminating motion and video compression, fluorescent light with flicker is highly likely to have an adverse effect on accuracy fluctuations, and eliminating flicker is highly likely to improve the results from the DNN model.

Figure~\ref{fig:efficientdet_flicker} shows the object detection counts for Efficient{D}et-v0 when there is no motion, video compression or flickering light. We observe fluctuations in object detection count up to 13.0\% (\textit{F2}) and 14.0\% (\textit{F10}). Due to space reasons, we have not included the graphs for with and without compression for Efficient{D}et-v0. However, like the YOLOv5 case, eliminating motion, video compression and flickering light improves the detection results.

\begin{figure}[tp]
    \centering
    \begin{subfigure}[t]{0.32\textwidth}
        \vspace{-0.6\baselineskip}
        \centering
        \includegraphics[width=0.99\textwidth]{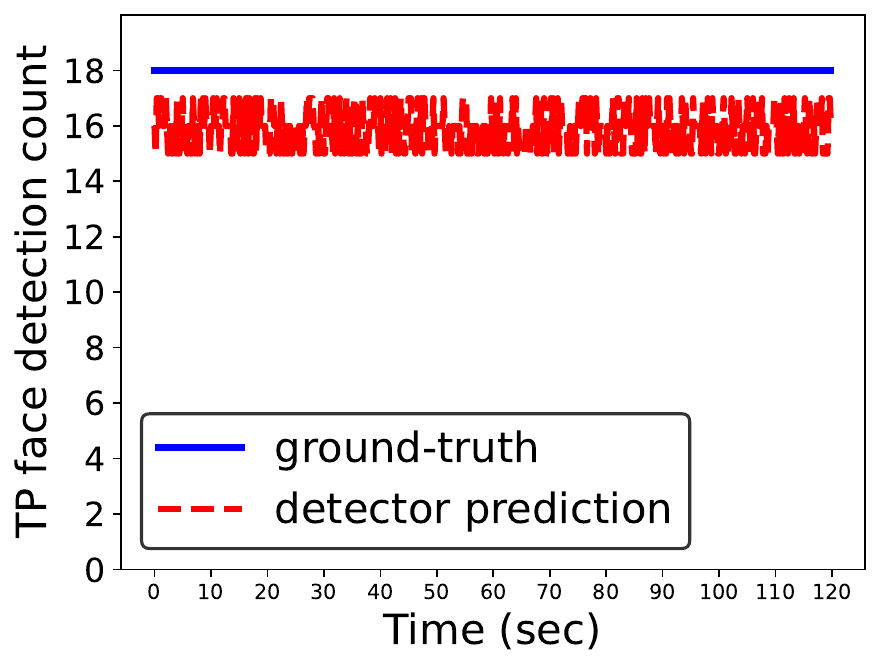}
        \caption{RetinaNet (Resnet-50)}
        \label{fig:retinaNet_flicker}
    \end{subfigure}
     \hfill
    \begin{subfigure}[t]{0.32\textwidth}
        \vspace{-0.6\baselineskip}
        \centering
        \includegraphics[width=0.99\textwidth]{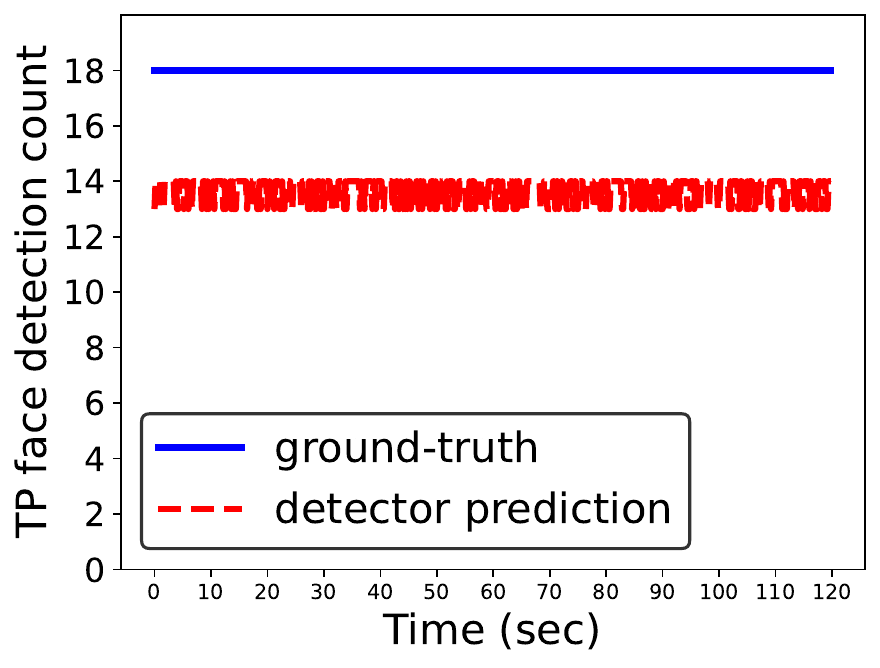}
        \caption{RetinaNet (Mobilenet)}
        \label{fig:mobilenetNet_flicker}
    \end{subfigure}
    \hfill
    \begin{subfigure}[t]{0.32\textwidth}
        \vspace{-0.6\baselineskip}
        \centering
        \includegraphics[width=0.99\textwidth]{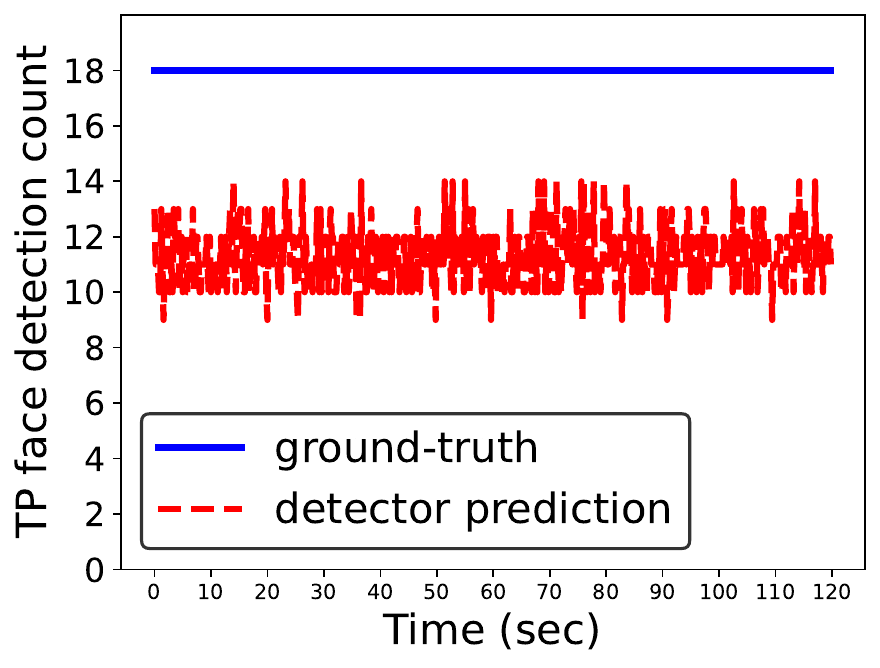}
        \caption{MTCNN}
        \label{fig:neoface_flicker}
    \end{subfigure}
    \vspace{-0.1in}
\caption{Face detection counts for three different DNN models when there is no motion, video compression or flickering light.}
\label{fig:face_flicker}
\vspace{-0.1in}
\end{figure}

Our detailed analysis in this section shows that eliminating motion, video compression and flicker does improve the object detection results. However, 
even after controlling for motion, video compression and flickering light, noticeable fluctuations in object detection counts still remain. We repeated the above experiments for three SOTA open-source \textit{face detection models.}
Fig. \ref{fig:face_flicker} shows fluctuation in face detection counts when there is no motion, video compression or flickering light. \textit{F2} metric reports true-positive face detection fluctuations up to 8.7\%, 4.3\%, 21.7\% for two retinaNet models and MTCNN respectively. 

\vspace{-0.15in}
\subsection{Impact of different camera models}
\label{subsec:diff-models}
\vspace{-0.1in}
\begin{figure}
    \centering
    \begin{subfigure}[t]{0.475\textwidth}
        \vspace{-0.6\baselineskip}
        \centering
        \includegraphics[width=0.99\textwidth]{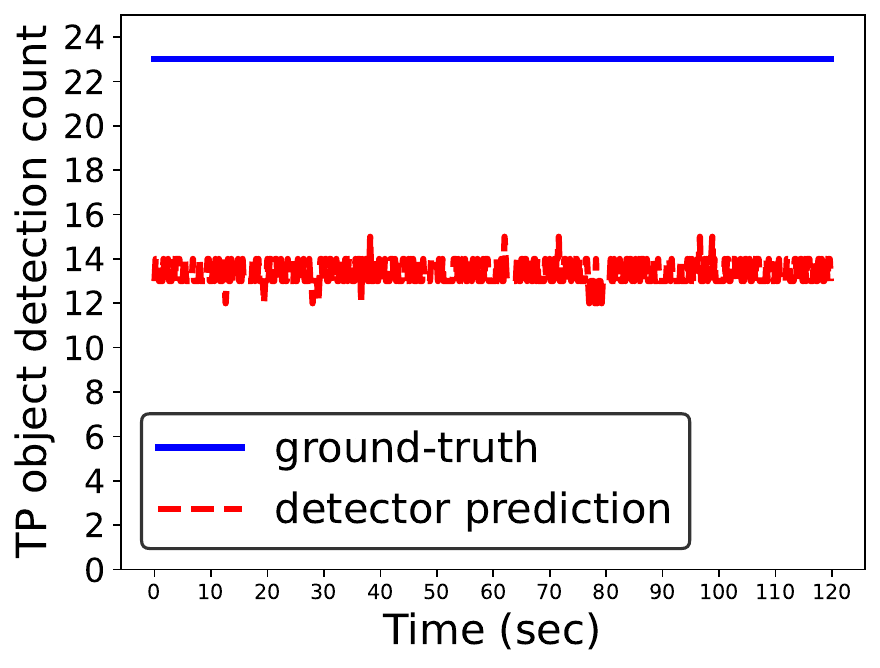}
        \caption{AXIS Q3515}
        \label{fig:camera_2_flicker}
    \end{subfigure}
    \hfill
    \begin{subfigure}[t]{0.475\textwidth}
        \vspace{-0.6\baselineskip}
        \centering
        \includegraphics[width=0.99\textwidth]{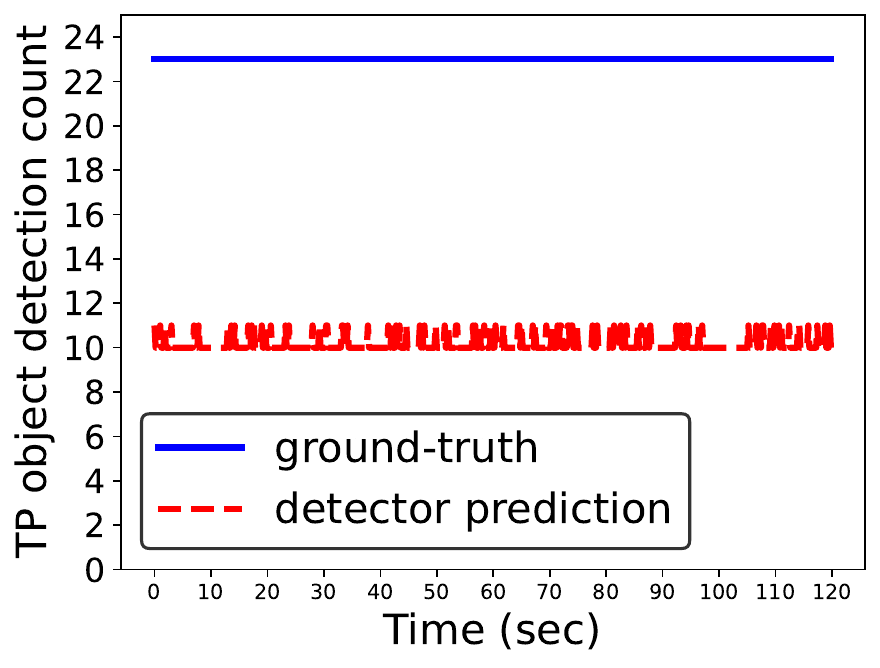}
        \caption{PELCO SARIX IME322}
        \label{fig:camera_3_flicker}
    \end{subfigure}
    \vspace{-0.1in}
\caption{Performance of YOLOv5
on different IP camera models in absence of motion, compression and flicker in lighting.}
\label{fig:cam_obj_flicker}
\vspace{-0.3in}
\end{figure}
We also investigated whether the fluctuation in video
analytics accuracy is observed only on specific camera model or is it
present across different camera models across different
vendors. Fig. \ref{fig:cam_obj_flicker} shows the performance of
{YOLO}v5 object detection model on {AXIS Q3515} and {PELCO} SARIX
IME322 IP cameras, both of them observing the same scene and in
absence of motion, compression and flicker in lighting. We note that
both of them show fluctuation in the count of detected objects and \textit{F2} metric reports up to 13.1\% and 4.4\% fluctuations for the two camera models. This
shows that the fluctuation in video analytics accuracy is observed
across different camera models from different vendors.
\vspace{-0.2in}
\section{Camera as an unintentional adversary}
\label{sec:understanding}
\vspace{-0.1in}


In~\secref{sec:investigation}, we investigated
various factors external to the camera
that could lead to fluctuations in video analytics
accuracy. Specifically, we looked at motion, compression, flicker in
lighting, and camera models from different vendors and different deep
learning models, and found out that fluctuation is observed across
different deep learning models, on different cameras, even when motion, compression, flicker in lighting are eliminated.
This leads us to hypothesize that the remaining factors causing accuracy fluctuation
may not be external.
Rather, it could 
be {\em internal} to the camera.
\vspace{-0.1in}
\subsection{Hypothesis}
\vspace{-0.05in}
\paragraph{Auto-parameter setting in modern cameras.}
Along with exposing endpoints to retrieve streaming videos
(\eg {RTSP} stream URL), IP cameras also expose {API}s to set various
camera parameters (\eg VAPIX~\cite{vapix} API for Axis camera
models). These camera settings aid in changing the quality of image
produced by the camera. Camera vendors expose these APIs because they
do not know ahead of time in what environment their camera would be
deployed and what settings would be ideal for that
environment. Therefore, they set the camera settings to some default
values and let end users decide what settings would work best for
their environment. There are two types of camera settings that are
exposed by camera vendors:
(1) Type 1 parameters include those that affect the way raw images are captured,
\eg exposure, gain, and shutter speed. These parameters generally are adjusted
{\em automatically} by the camera with little control by end users. They
only allow end users to set maximum value, but within this value, the
camera internally changes the settings dynamically in order to produce a visually
pleasing video output. We refer to such parameters as automated
parameters ({AUTO}).
(2) Type 2 parameters include those that affect processing of raw data in order to
produce the final frame, \eg image specific parameters
such as brightness, contrast,
sharpness, color saturation, and video specific parameters such as
compression, GOP length, target bitrate, FPS. For these parameters, camera vendors
often expose fine control to end users, who can set the specific value.
We refer to such parameters as non-automated parameters
({NAUTO}). 
The distinction between {AUTO} and {NAUTO} parameters help us refine
our hypothesis where we can fix values of {NAUTO} parameters, vary the
maximum value for {AUTO} parameters and observe how camera internally
changes these parameters to produce different consecutive frames,
which might be causing the fluctuations. 


\textbf{The Hypothesis.}
The purpose of a video camera is to capture videos, rather than still
images, for viewing by human eyes. Hence, irrespective of how the
scene in front of the camera looks like, \ie whether the scene is
static or dynamic, video camera always tries to capture a video, which
assumes changes in successive frames. To capture a visually pleasing and smooth 
(to human eyes) video,
the camera tries to find the optimal exposure time or shutter
speed. 
On one hand, high shutter speed, \ie low
exposure time, freezes motion in each frame, which results in very
crisp individual images. However, when such frames are played back at
usual video frame rates, it can appear as hyper-realistic and provide
a very jittery, unsettled feeling to the viewer
\cite{understanding-video}.
Low shutter speed, on the other hand, can cause moving objects to appear
blurred and also builds up noise in the capture. To maintain
appropriate amount of motion blur and noise in the capture, video
cameras have another setting called \emph{gain}. Gain indicates the
amount of amplification applied to the capture. A high gain can
provide better images in low-light scenario but can also increase the
noise present in the capture. 
For these reasons, the optimal values of {AUTO}
parameters like exposure and gain are internally adjusted by the camera
to output a visually pleasing smooth video.
Thus, video capture is fundamentally different from still image
capture and the exact values of exposure and gain used by the camera
for each frame are not known to end users or analytics applications
running on the video output from the camera. This loose control over
maximum shutter time and maximum gain parameters is likely the
explanation for fluctuations in video analytics accuracy,
\ie the camera's unintentional adversarial effect.

\vspace{-0.1in}
\subsection{Hypothesis validation}
\label{subsec:validation}
\vspace{-0.1in}
The above explanation of our hypothesis that internal dynamic change
of {AUTO} parameters applied by a camera causes successive frames
to differ and hence fluctuations in video analytics accuracy, also
points us a way to partially validate the hypothesis.
\begin{figure}
    \centering
    \begin{subfigure}[t]{0.485\textwidth}
    \vspace{-0.7\baselineskip}
        \centering
        \includegraphics[width=0.99\textwidth]{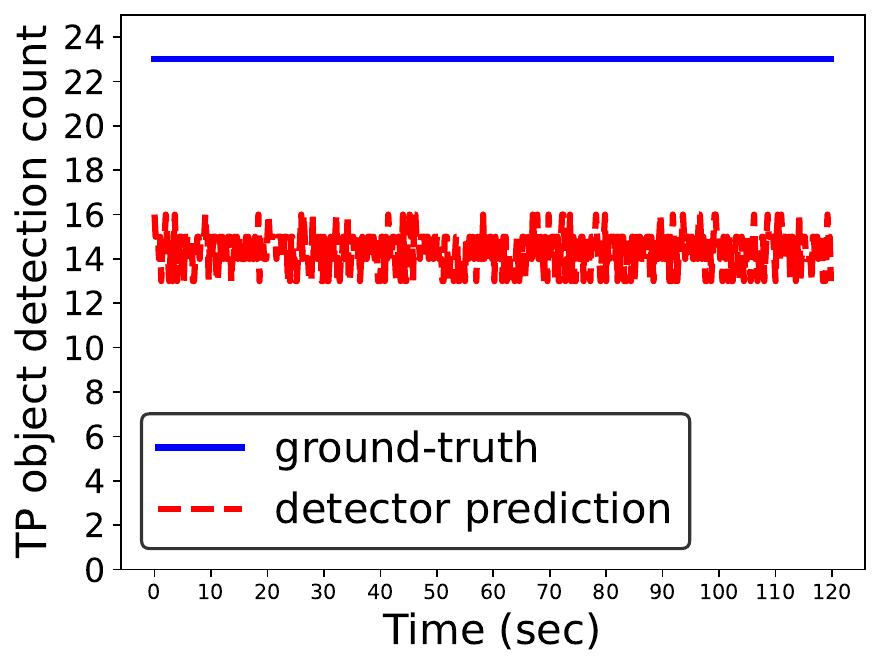}
        \caption{Max exposure time = 1/4 second}
        \label{fig:obj_exposure2}
    \end{subfigure}
    \hfill
    \begin{subfigure}[t]{0.485\textwidth}
    \vspace{-0.7\baselineskip}
        \centering
        \includegraphics[width=0.99\textwidth]{figs/camera_1_yolov5_exposure_1_flicker_free_count.pdf}
        \caption{Max exposure time = 1/120 second}
        \label{fig:obj_exposure1}
    \end{subfigure}
    \vspace{-0.1in}
\caption{Performance of YOLOv5 Object detectors for two different settings of an AUTO parameter, in absence of motion, compression and flicker in lighting.}
\label{fig:exposure_yolov5}
\vspace{-0.25in}
\end{figure}
Since the camera still exposes control of maximum values of {AUTO} parameters, we can adjust these maximum parameter value 
and observe the impact on the resulting fluctuation of analytics accuracy.
Fig. \ref{fig:exposure_yolov5} shows the fluctuation in
accuracy of {YOLO}v5 object detection model for two different settings
of maximum exposure time. 
We observe that when the maximum exposure time is 1/120 second, the
fluctuation in object count is significantly lower than when it is 1/4
second. Here, reducing the max exposure time decreases the maximum \textit{F2} fluctuations from 13.0\% to 8.7\%. This corroborates our hypothesis -- 
with a higher value of maximum exposure time,
the camera can possibly choose from a larger number of different exposure
times than when the value of maximum exposure time is low, which in
turn causes the consecutive frame captures to differ more, resulting
in more accuracy fluctuation. 

We compared the sequences of object detection counts  at a maximum exposure of 1/120 seconds and 1/4 seconds using the t-test for repeated measures, and  easily rejected the null hypothesis that lowering the maximum exposure time (\ie changing exposure time from 1/4 seconds to 1/120 seconds) makes no difference in detection counts, at a 0.01 level of significance (99\% confidence). Therefore, the choice of maximum exposure time has a direct impact on the accuracy of the deep learning models, and the precise exposure time is automatically determined by the video camera. We quantify this using object tracking task (discussed in Section \ref{sec:implications}) and observe 65.7\% fewer mistakes in tracking when exposure changes.



%


\vspace{-0.15in}
\section{Implications}
\label{sec:implications}
\vspace{-0.05in}
SOTA object detectors (Yolov5~\cite{glenn_jocher_2022_6222936} or
Efficient{D}et~\cite{tan2020efficientdet}) are trained on still image
datasets~\eg COCO~\cite{lin2014microsoft} and VOC~\cite{pascal-voc}
datasets. We observed in prior sections that the accuracy of insights from deep learning models fluctuate when 
used for video analytics tasks. 
An immediate
implication of our finding is that 
deep learning models trained on still image datasets should not be directly
used for videos. We discuss several avenues of research that can mitigate the accuracy fluctuations in video analytics tasks due to the use of image-trained 
DNN models. 
\vspace{-0.15in}
\subsection{Retraining image-based models with transfer learning} 
One relatively straight-forward approach is to train models for extracting insights from videos
using video frames that are captured under different scenarios.
As a case study,
we used transfer learning to train Yolov5 using the proprietary videos captured under different scenarios. These proprietary videos contain objects from person and vehicle super-category (that have car, truck, bus, train categories), captured by the cameras at different deployment sites (\eg traffic intersection, airport, mall, etc.) during different times-of-the-day (\ie day, afternoon, night) and also under different weather conditions (\ie rainy, foggy, sunny). We extract total \emph{34K} consecutive frames from these proprietary video snippets, and these frames form our training dataset.

\textbf{Training details.} The first 23 modules (corresponding to 23 layers) of our new deep learning model are initialized
using weights from COCO-trained Yolov5 model, and these weights are frozen. During
training, only the weights in the last \textit{detect} module are
updated. For transfer learning, we used a learning rate of 0.01 with a
weight decay value of 0.0005. We trained Yolov5 model on NVIDIA GeForce
RTX 2070 GPU server for 50 epochs with a batch size of 32. This
lightweight training required only 1.6 GB GPU memory and took less than
1.5 hours to finish 50 epochs. We used the newly trained Yolov5 model to detect objects (\ie cars and persons) in (a) our static scene of 3D models, and (b) a video from the Roadway dataset (same video that was  used in~\secref{sec:motivation}). 

\begin{figure}[tb]
    \centering
    \begin{subfigure}[t]{0.495\textwidth}
        \vspace{-0.7\baselineskip}
        \centering
        \includegraphics[width=0.99\textwidth]{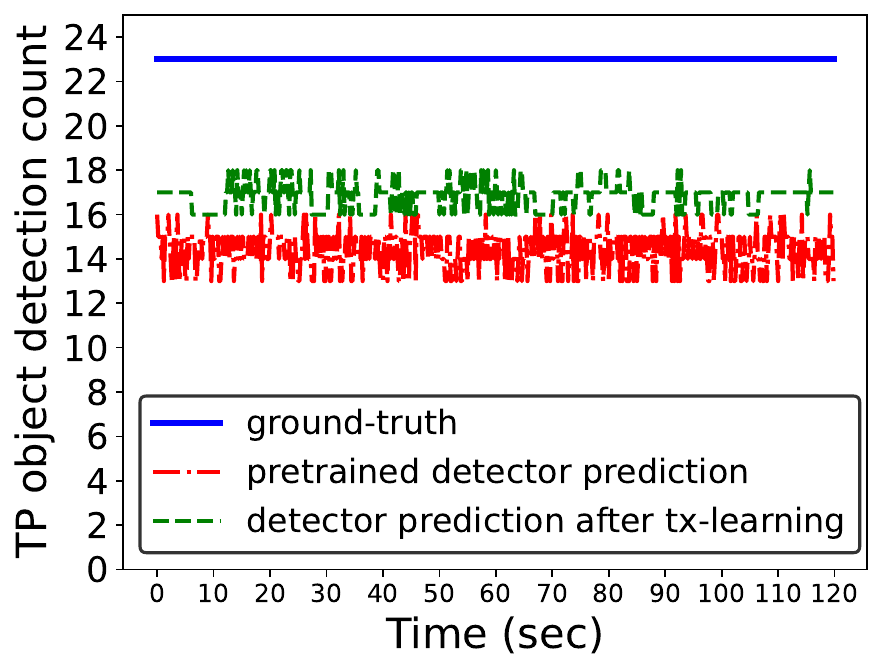}
        \caption{Static Scene of 3D models}
        \label{fig:static_final}
    \end{subfigure}
    \hfill
    \begin{subfigure}[t]{0.495\textwidth}
        \vspace{-0.7\baselineskip}
        \centering
        \includegraphics[width=0.99\textwidth]{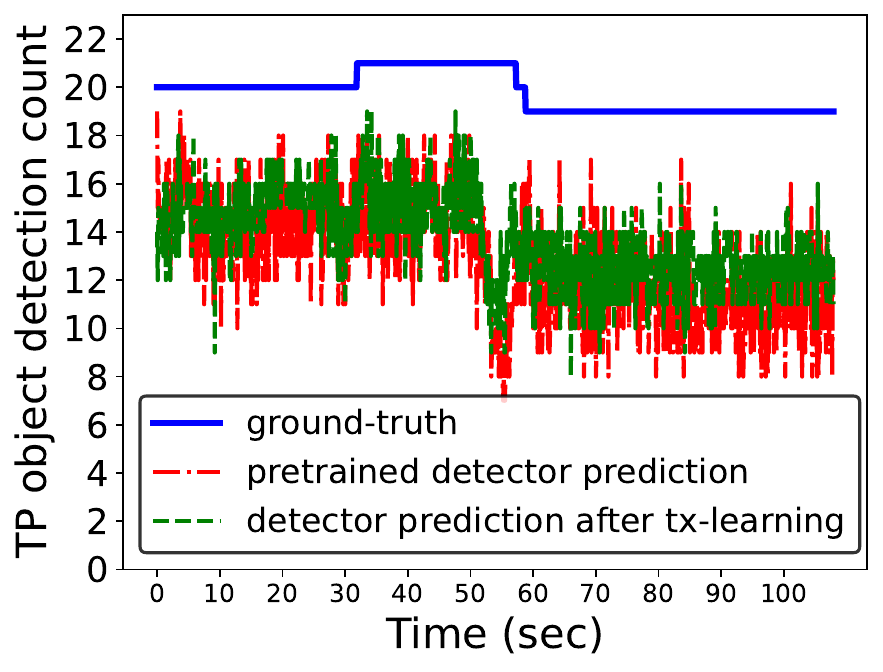}
        \caption{Video snippet from Roadway dataset}
        \label{fig:roadway_1_final}
    \end{subfigure}
    \vspace{-0.1in}
\caption{Detection counts from YOLOv5-large 
after transfer-learning}
\label{fig:yolov5-final}
\vspace{-0.25in}
\end{figure}
\figref{fig:static_final} shows the improvement in detection counts due to the transfer-learning trained Yolo5 model (green waveform). The improvement over the original Yolo5 model (shown as red waveform) is noticeable visually. 
We also compared the sequence of object detection counts for the original Yolov5 model (red waveform) and the transfer-learning trained Yolov5 model (green waveform) by using a t-test for repeated measures.  We easily rejected the null hypothesis that transfer-learning  makes no difference, at a 0.01 level of significance (99\% confidence). Then, we estimated the size of the effect due to transfer-learning, and we observed that at a 0.01 level of significance, the improvement is 2.32 additional object detections (14.3\% improvement over the mean detections due to the original Yolov5 model). For this experiment, the camera was automatically setting AUTO camera parameters to produce a visually pleasing video, and the transfer-learning trained Yolov5 detector was able to detect more objects despite the unintentional adversary (camera). 

In practical deployments of video analytics systems that operate 24x7, it is difficult to control motion or environmental conditions, and the default video compression settings also vary from camera to camera. To understand the impact of transfer-learning trained Yolov5 model, we did experiments on videos in the Roadway dataset. These videos have motion, and the environmental conditions or video compression settings are unknown (such information is not part of the Roadway dataset).
\figref{fig:roadway_1_final} shows the results for a video in the Roadway dataset.  The true-positive object detections by our \emph{transfer-learning trained Yolov5 model} (green waveform) show noticeably less range of fluctuations than the original Yolov5 model (red waveform).
We also compared the sequence of object detection counts for the original Yolov5 model (red waveform) and the transfer-learning trained Yolov5 model (green waveform) by using a t-test for repeated measures.  We easily rejected the null hypothesis that transfer-learning  makes no difference, at a 0.01 level of significance (99\% confidence). Then, we estimated the size of the effect due to transfer-learning, and we observed that at a 0.01 level of significance, the improvement is 1  additional object detection (9.6\% improvement over the mean detections due to the original Yolov5 model).
Our \emph{newly trained} Yolov5 model reduces the maximum variation of correctly detected object counts from 47.4\% to 33.2\% \textit{(F10)}, and 42.1\% to 32.5\% \textit{(F2)}. 


\textbf{Impact on object tracking.} We evaluated the impact of the fluctuations in detection counts on object tracking task where we track the objects across different frames using \emph{MOT SORT}~\cite{Bewley2016_sort} tracker.
Object trackers assign the same track-id to an object appearing in contiguous frames. If an object is not detected in a frame, then the object's track is terminated. If the object is detected again in subsequent frames, a new track-id is assigned to the object. We use the number of track-ids assigned by a tracker as an indicator of the quality of object detections.
Our tracker reported \emph{157} track-ids when the original Yolov5 model was used for detecting objects in the video from the Roadway dataset. In contrast, the same tracker reported \emph{94} track-ids when the \emph{transfer-learning trained} Yolov5 model was used (\ie 40.1\% fewer mistakes in tracking). We manually annotated the video and determined the ground-truth to be 29 tracks. We also manually inspected the tracks proposed by the tracker for the two models (with and without transfer-learning) to ensure that the tracks were true-positives. These experiments suggest that the transfer-learning based Yolov5 model leads to better performance on object tracking task. 

\vspace{-0.1in}
\subsection{Calibrating softmax confidence scores}
\vspace{-0.02in}
In general, we use softmax confidence output as the correctness probability estimate of the prediction. However, many of these neural networks are poorly calibrated~\cite{guo2017calibration,minderer2021revisiting}. 
The uncertainty in softmax confidence scores from poorly calibrated NN can potentially worsen the robustness of video analytics performance. To mitigate this, we can employ several {post-hoc} methods on SOTA models to improve the softmax estimates, \eg via averaging predictions obtained from bag of models (\eg detectors, classifiers)~\cite{lakshminarayanan2017simple}, platt scaling~\cite{guo2017calibration}, isotonic regression~\cite{nyberg2021reliably}, etc. We can also adapt the confidence threshold 
to filter out the low-confidence mispredictions.
This confidence threshold value can be adapted based on the difficulty level to detect in a certain frame. We leave the investigation of neural network calibration and confidence threshold adaptation as future work.


\vspace{-0.15in}
\section{Related work}
\label{sec:related}

Several
efforts~\cite{najafi2019robustness,zhang2019theoretically,cheng2020cat,jin2020manifold,madry2018towards}
have been made to improve the robustness of deep learning models against
adversarial attacks.
Recent works~\cite{Xie_2020_CVPR,Hendrycks_2021_CVPR} propose several
adversarial examples that can be used to train a robust model and also
serve as performance measures under several distribution
shifts. Robust model training based on shape representation rather
than texture-based representation is proposed
in~\cite{geirhos2018imagenet}. Xie et. al. ~\cite{Xie_noisy_2020_CVPR}
use unlabeled data to train SOTA model through \emph{noisy student
  self distillation} which improves the robustness of existing
models. However, the creation of these ``robust" models does not take
into account the kind of adversaries introduced by dynamic tuning of
AUTO parameters by video cameras.  Also, the perturbations introduced
in variants of {I}mage{N}et dataset (\ie -C, -3C, -P
\etc)~\cite{hendrycks2019robustness} are not the same as observed when
    {AUTO} parameters are tuned, which makes such datasets unsuitable
    for our study.

While there have been many
efforts~\cite{chen2015glimpse,du2020server,jiang2018chameleon,kang2017noscope,li2020reducto,liu2019edge,zhang2018awstream,zhang2017live}
on saving compute and network resource usage without impacting the accuracy
of video analytics pipelines (VSPs) by adapting different video-specific
parameters like frame rate, resolution, compression, \etc, there has
been little focus on improving the \textit{accuracy} of VAPs. Techniques to improve the accuracy of VSPs by
dynamically tuning camera parameters is proposed by Paul
et. al~\cite{camtuner}, but they focus on image-specific NAUTO
parameters rather than AUTO parameters, which we show is the cause for
``unintentional" adversarial effect introduced by the camera. Otani
et. al.~\cite{otani2022performance} show the impact of low video quality
on analytics, but they study network variability rather than the
camera being the reason for low video quality.





Koh et. al. ~\cite{pmlr-v139-koh21a} identify the need for training models with the distribution shift that will be observed in practice in real-world deployment. 
Inspired from this, rather than using independent images or synthetically transformed images for training, we use real video frames for training, which takes into account the distribution shift observed in practice for video analytics.

Wenkel et. al~\cite{wenkel2021confidence} tackle the problem of
finding optimal confidence threshold for different models in model training
and mention
the challenge that there is a possibility of fluctuation in accuracy
based on small changes between consecutive frames. However, they do
not go in depth to analyze it further as we do.
To our best knowledge, we are the first to expose the camera as an ``unintentional adversary" for video analytics task and propose 
mitigation techniques.

\vspace{-0.15in}

\section{Conclusion}
\vspace{-0.1in} 

In this paper, we show that blindly applying
image-trained deep learning models for video analytics tasks leads to
fluctuation in accuracy. We systematically eliminate 
external
factors including motion, compression and environmental conditions
(e.g., lighting) as possible reasons for fluctuation and show that the
fluctuation is due to internal parameter changes applied by the camera,
which acts as an ``unintentional adversary" for video analytics
applications. To mitigate this adversarial effect, we propose
a transfer learning based approach and train a new Yolov5 model for
object detection. We show that by reducing fluctuation across frames,
our model is able better track objects ($\sim$ 40\% fewer mistakes in
tracking). Our paper exposes a fundamental fallacy in applying deep
learning models for video analytics and opens up new avenues for
research in this direction.


\vspace{-0.1in} 
\paragraph{Acknowledgment.}
This  project is supported in part 
by NEC Labs America
and by NSF grant 2211459.

\bibliographystyle{splncs04}
\bibliography{reference}

\end{document}